\documentclass[times, 10pt,twocolumn]{article} 
\usepackage{latex8}
\usepackage{times}
\usepackage{booktabs}
\usepackage[flushleft]{threeparttable}
\usepackage{graphicx}
\usepackage{dblfloatfix}

\begin{document}

\title{Swin-Pose: Swin Transformer Based Human Pose Estimation}

\author{Zinan Xiong\thanks{$^{\ast}$Equal contribution}\\
Department of Computer Science\\ Universify of Massachusetts Lowell\\ Lowell, MA 01851, USA\\ zinan\_xiong@student.uml.edu\\
\and
Chenxi Wang$^{\ast}$\\
Department of Electrical \& Computer Engineering\\ Universify of Massachusetts Lowell\\ Lowell, MA 01851, USA\\ chenxi\_wang1@student.uml.edu
\and
Ying Li$^{\ast}$\\
Alibaba Group, Hangzhou, China\\ yunle.ly@alibaba-inc.com
}

\providecommand{\keywords}[1]{\textbf{\textit{Index terms---}} #1}

\maketitle
\thispagestyle{empty}

\begin{abstract}
Convolutional neural networks (CNNs) have been widely utilized in many computer vision tasks. However, CNNs have a fixed reception field and lack the ability of long-range perception, which is crucial to human pose estimation. Transformer architecture has been adopted to computer vision applications recently and is proven to be a highly effective architecture. We are interested in exploring its capability in human pose estimation, and thus propose a novel model based on transformer, enhanced with a feature pyramid fusion structure. More specifically, we use pre-trained Swin Transformer to extract features, and leverage a feature pyramid structure to extract and fuse feature maps from different stages. The experiment results of our study have demonstrated that the proposed transformer-based model can achieve better performance compared to the state-of-the-art CNN-based models. 
\end{abstract}
\vspace{-0.2in}


\begin{figure*}
    \centering
    \includegraphics[width=0.8\textwidth]{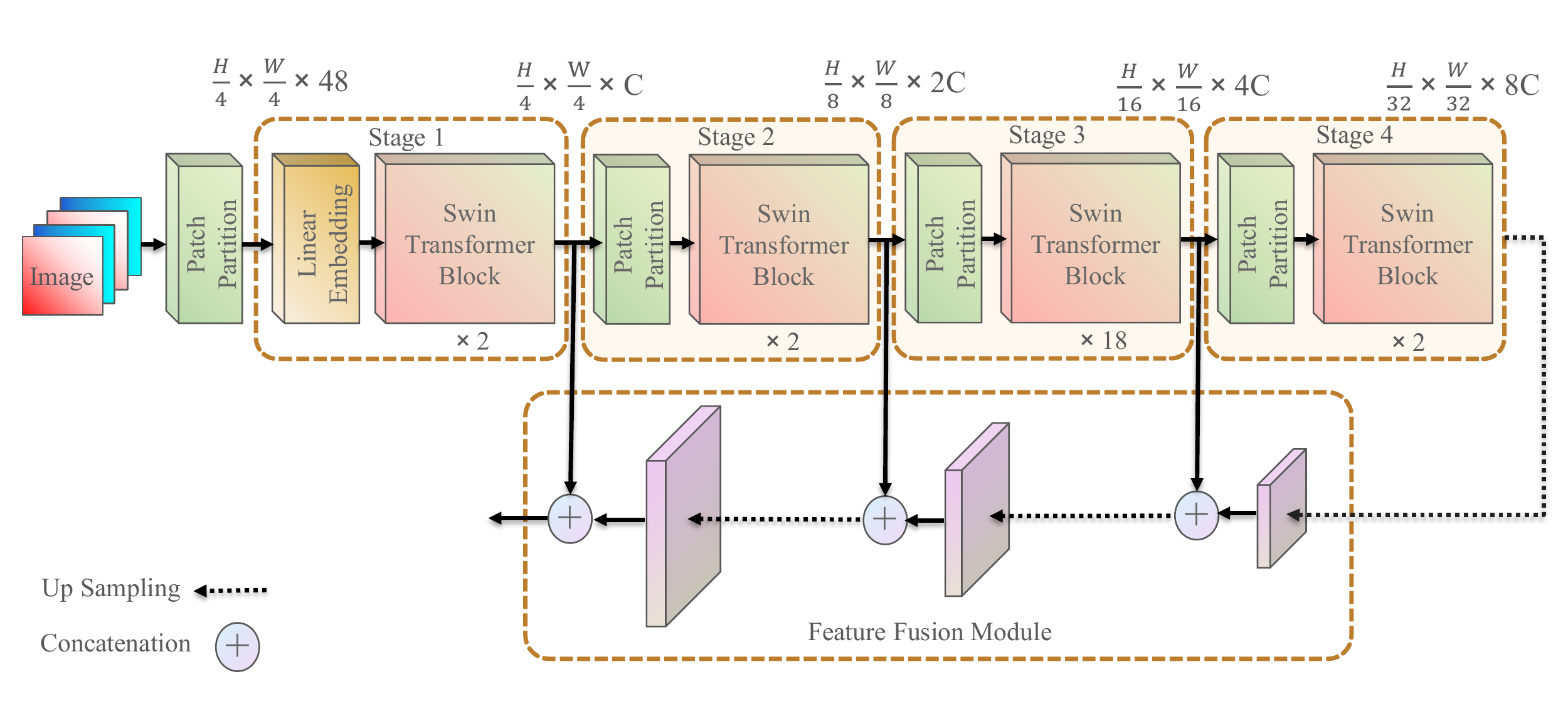}
    \caption{Swin-Pose Architecture Overview.}
    \label{fig:overall_structure}
    \vspace{-0.2in}
\end{figure*}

\Section{Introduction}
\vspace{-0.1in}
Human pose estimation is one of the key tasks in computer vision, it aims to detect and locate persons with their body joints such as neck and shoulder. It has several important and promising applications, including action recognition, human-computer interaction, and gaming. There are still a lot of challenges brought by partial or full joint occlusions, differences in body shapes, and clothing. A large amount of work is devoted to obtain a better feature representation and distinguish the correct pose. However, the computation cost is high and its ability to adapt to all poses is still limited.

Recently, deep convolutional neural networks (DCNNs) has proved its ability in computer vision tasks. Among all these approaches, two of them are mainstream methods: keypoint position regression\cite{toshev2014deeppose}, and keypoint heatmap estimation followed by choosing the location with the highest score\cite{chu2016structured,li2020human,wang2021lower}. The former one treats pose estimation as a joint position regression problem. The latter one puts a 2D Gaussian kernel on all keypoints, constructs ground truth heatmaps, and utilizes these heatmaps to supervise the prediction with L2 loss. The heatmap regression is relatively simpler to implement than the approach of keypoint regression, and it achieves both higher accuracy and efficiency.

The attention mechanism calculates the similarity and weights between the current input and all the other inputs, then the model will put more effort into the context which has a higher weight.
Transformer adopts the structure of encoder-decoder, but it has demonstrated that the replacement of the original design with multi-head self-attention leads to a better performance\cite{vaswani2017attention}. 
However, traditional transformer structure is unable to be used on the human pose estimation task directly. Its computation cost is tremendously high, and will be increased sharply with the size of feature map and consuming excessive memory. In addition, traditional transformer only generates the output feature map within one single scale, thus it has a limit performance on images with multi-scale objects.
       
To leverage the long-range dependencies capturing ability of transformer, and avoid the excessive memory consumption, we choose Swin Transformer \cite{liu2021swin} as our backbone. We also integrate it with a top-down feature pyramid to add the ability of scale-invariant. With this Swin Transformer structure and two different fusion methods, we propose a novel human pose estimation model Swin-Pose, it achieves outstanding results compared with HRNet-W48. Additionally, we perform a number of experiments and investigate the effects of input resolution, model size, and fusion method, and present the results in Section \ref{ablationstudy}.

\Section{Related Work}
\vspace{-0.1in}
\SubSection{Human Pose Estimation}
\vspace{-0.1in}
CNN has demonstrated its powerful ability in computer vision, owing to its characteristics of local connectivity, and weight sharing, thus also being widely applied to human pose estimation tasks. 
Convolutional Pose Machine generates 2D belief maps for the location of each part by several convolutional networks at each stage\cite{wei2016convolutional}, which encodes the spatial uncertainty of the locations, and captures the spatial relationships between the parts. Simple Baseline\cite{xiao2018simple} uses ResNet\cite{he2016deep} as the backbone to extract features, three deconvolutional layers are added to the last convolution stage, and generate the feature maps and heatmaps with high resolution. 

Instead of following one route, HRNet\cite{wang2020deep} recovers the high-resolution from low-resolution at sub-network and maintains high-resolution through the whole process. It gradually adds branches with lower resolution in parallel, but fuses the output from all branches, then passes the fused information through all branches, and finally generates the keypoint estimation.

\begin{figure}[h]
    \centering
    \includegraphics[width=0.8\columnwidth]{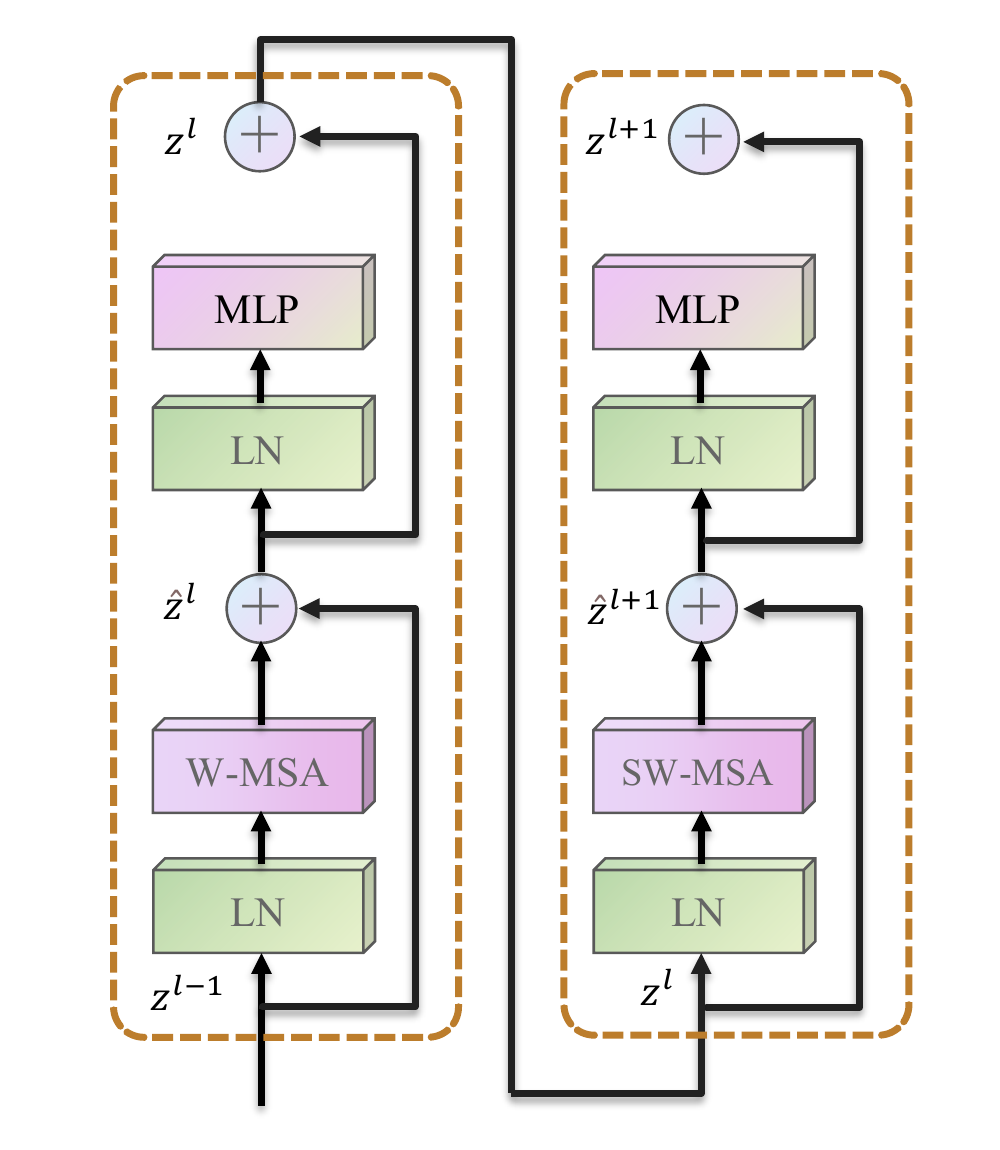}
    \caption{Structure of Transformer Block.}
    \label{fig:swin-transformer-block}
    \vspace{-0.2in}
\end{figure}

\SubSection{Attention Mechanism and Transformer}
\vspace{-0.1in}
Originally, transformer models were applied to natural language processing (NLP). It stacks encoder and decoder together, both of which are consist of self-attention and feed-forward layers. The self-attention layer maps query, key and value to the output then computes the output as a weighted sum of values. A multi-head mechanism is also implemented to obtain the attention between different positions jointly and achieves a remarkable result\cite{vaswani2017attention}.


Vision Transformer (ViT) splits the input images into patches of fixed size, then applies linear embeddings to each patch and send them through the transformer as tokens\cite{dosovitskiy2020image}. A learnable classification embedding is prepended to the patch, and a position embedding is also added to keep the positional information. 
TransPose is proposed to combine the translation equivalence of CNN and the long-range dependencies capturing ability of the transformer\cite{yang2021transpose}. It uses convolutional neural network as its backbone, then connect to a transformer encode layer and captures the global dependencies at high-level feature maps. 

The structure of Swin Transformer is similar to ResNet and consists of four stages. By limiting the self-attention within non-overlapping local windows, Swin Transformer reduces the computation cost by a large margin\cite{liu2021swin}, making it applicable in downstream tasks. A shifted window partitioning approach is also applied to enable information communication between those non-overlapping windows. 


\SubSection{Image Feature Fusion}
\vspace{-0.1in}
Previous studies of feature fusion methods have established that multi-level features can be extracted and fused \cite{deng2020feature}. HyperNet \cite{kong2016hypernet} concatenate the representatives before prediction, Spatial Pyramid Pooling \cite{he2015spatial} investigated the multi-scale filters. Furthermore, the image pyramid is famous for its ability to combine and utilize the features extracted from different scales and its characteristics of scale-invariant. Felzenszwalb \emph{et al.} \cite{felzenszwalb2009object} proposed an efficient and accurate system based on pictorial structures framework, they obtained the feature pyramids by repeated smoothing and sub-sampling from different levels, then a score at different positions and scales was identified. 

In 2017, Lin \emph{et al.} \cite{lin2017feature} described Feature Pyramid Networks for object detection. They identified a feature pyramid that has strong semantics from low to high levels to leverage the pyramidal shape of ConvNet's \cite{krizhevsky2012imagenet} feature hierarchy.

\section{Methodology}
\vspace{-0.1in}
We propose Swin-Pose model that utilises a multi-method approach combining the Swin Transformer Block and feature pyramid fusing. The overview of our proposed model architecture is shown in Fig. \ref{fig:overall_structure}, which follows the logic of the large version of Swin Transformer (Swin-L). The advantage of the swin transformer is that it allows us to extract important information and long-range dependencies between joints from the images. Before obtaining the embedding with the linear layer, the input images are split into four non-overlapping patches. After these patches are treated as ``tokens", the raw pixel RGB values are concatenated to produce the features. Once the embeddings are passed through four stages, feature maps are generated with different scales. 


Furthermore, the feature pyramid is one of the most practical ways of examining the inherent multi-scale features. It is applied to fuse the feature maps and produce the output for keypoint heatmap regression.

\begin{figure}
    \centering
    \includegraphics[width=0.9\columnwidth]{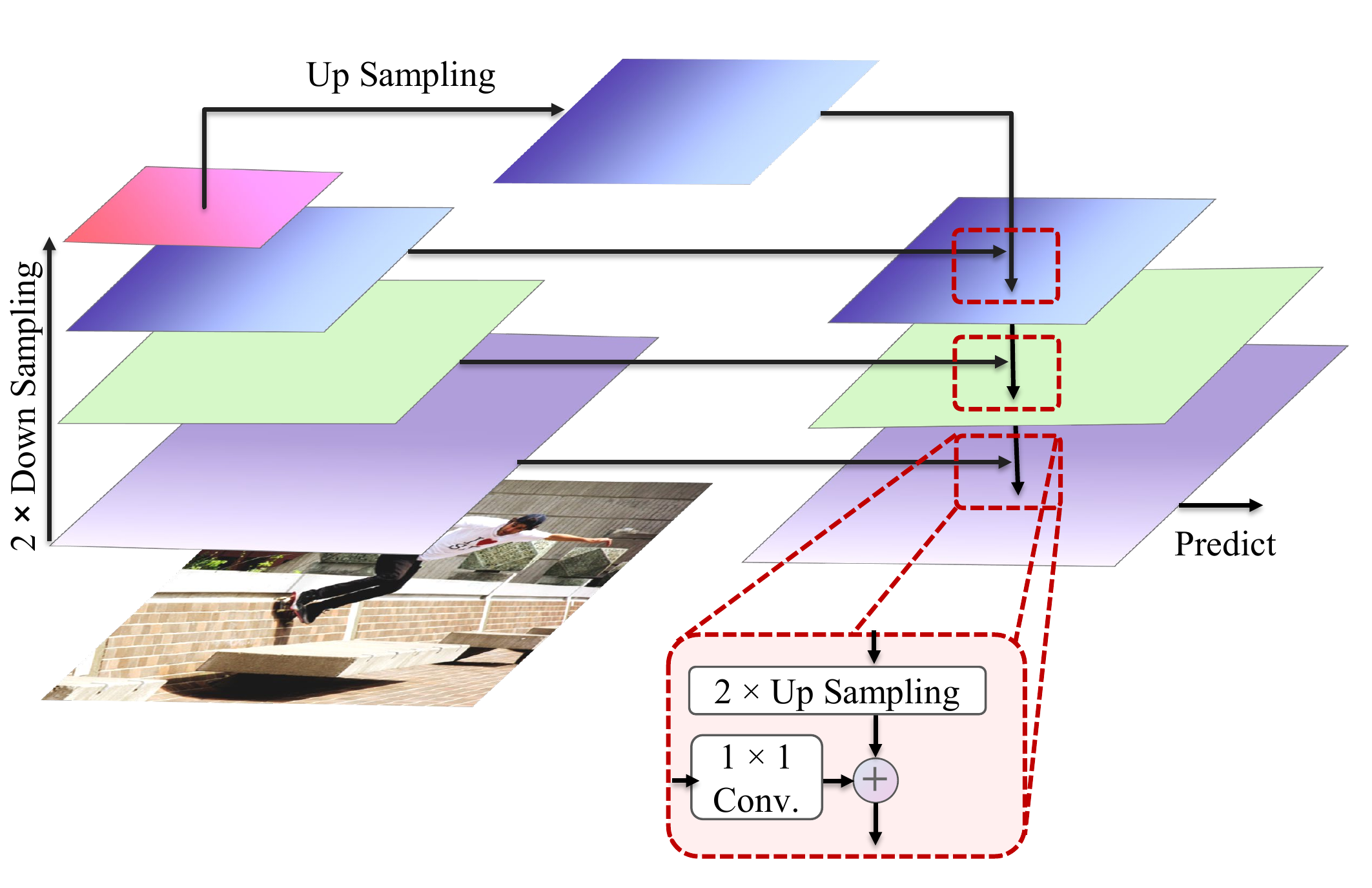}
    \caption{Feature Fusion Module.}
    \label{fig:fpn}
    \vspace{-0.2in}
\end{figure}

\subsection{Swin-Pose Transformer Module}
\vspace{-0.1in}
In our study, the vanilla Swin-Pose follows the settings in Swin Transformer, setting the patch size to $4 \times 4$. The tokens are sent to a linear embedding layer with a classification token prepended, and position information encoded, and its dimension is projected from $4 \times 4 \times 3 = 48$ to an arbitrary value $C$. A number of Swin Transformer blocks are applied to the embedded tokens and make up ``Stage 1" together with the aformentioned linear embedding layer. In order to get a hierarchical representation, the output of ``Stage 1" is then sent through a patch merging layer, where the $2 \times 2$ neighbouring patches are merged together. Also, the resolution is down-sampled by a factor of 2, and the number of tokens is reduced by a factor of 4. A linear layer is also applied to change the dimension from $4C$ to $2C$. As shown in Fig. 1, with the patch merging layer, multiple swin transformer blocks are connected, constituting three identical stages. Moreover, the resolutions of output are $\frac{H}{16} \times \frac{W}{16}$ and $\frac{H}{32} \times \frac{W}{32}$, respectively.

The proposed transformer block is derived from the original swin transformer block \cite{liu2021swin}. It consists of LayerNorm (LN) layer, multi-head self-attention layer, 2-layer multilayer perception with Gaussian Error Linear Unit (GELU), and residual connection structure. More importantly, the standard multi-head self-attention layer is replaced by window-based multi-head self-attention (W-MSA) and shifted window-based multi-head self-attention (SW-MSA), respectively, as shown in Fig. \ref{fig:swin-transformer-block}.

After extracting features from the vanilla Swin-Pose transformer backbone, a heatmap regression layer is applied to estimate the locations of all keypoints. 
This vanilla Swin-Pose structure was trained initially without the pre-trained model of ImageNet. Surprisingly, our experiments obtained weak $AP$ about 18\% lower than previously reported models. This result may be explained by the fact that transformer requires a large scale dataset to achieve competitive performance. Although swin transformer has powerful performance in extracting features from the connections of the patches from different locations, the human pose estimation task heavily depends on the local features near each keypoint. To improve the local feature extraction performance of the Swin-Pose Transformer Module, we propose a feature fusion module before the heatmap regression layer, whose details are explained in the following subsection. 
\subsection{Feature Fusion Module}
\vspace{-0.15in}
There are 4 stages in the structure of Swin Transformer, and the size of the output are $\frac{H}{4} \times \frac{W}{4}, \frac{H}{8} \times \frac{W}{8}, \frac{H}{16} \times \frac{W}{16}, \frac{H}{32} \times \frac{W}{32}$, respectively, where $H$ and $W$ are the height and width of the input image. 
In order to avoid situations where the model overfits to a specific scale and loses its generalization, we use top-down pathway and lateral connections to fuse the feature maps from different resolution levels and combine outputs of different resolutions and semantic information, as shown in Fig. \ref{fig:fpn}. During the fusion procedure, a $1 \times 1$ convolution is first used to change the channels of the output from the upper stage. Then a bi-linear up sampling is used to change the resolution while maintaining the number of channels, so both channels and resolution will match the output from the lower stage extracted through the lateral connection. Then we put them together, repeat the procedure of $1 \times $1 and up sampling twice, get the final output with the highest resolution, and use it for the next step of heatmap regression.

\begin{figure}
    \centering
    \includegraphics[width=0.8\columnwidth]{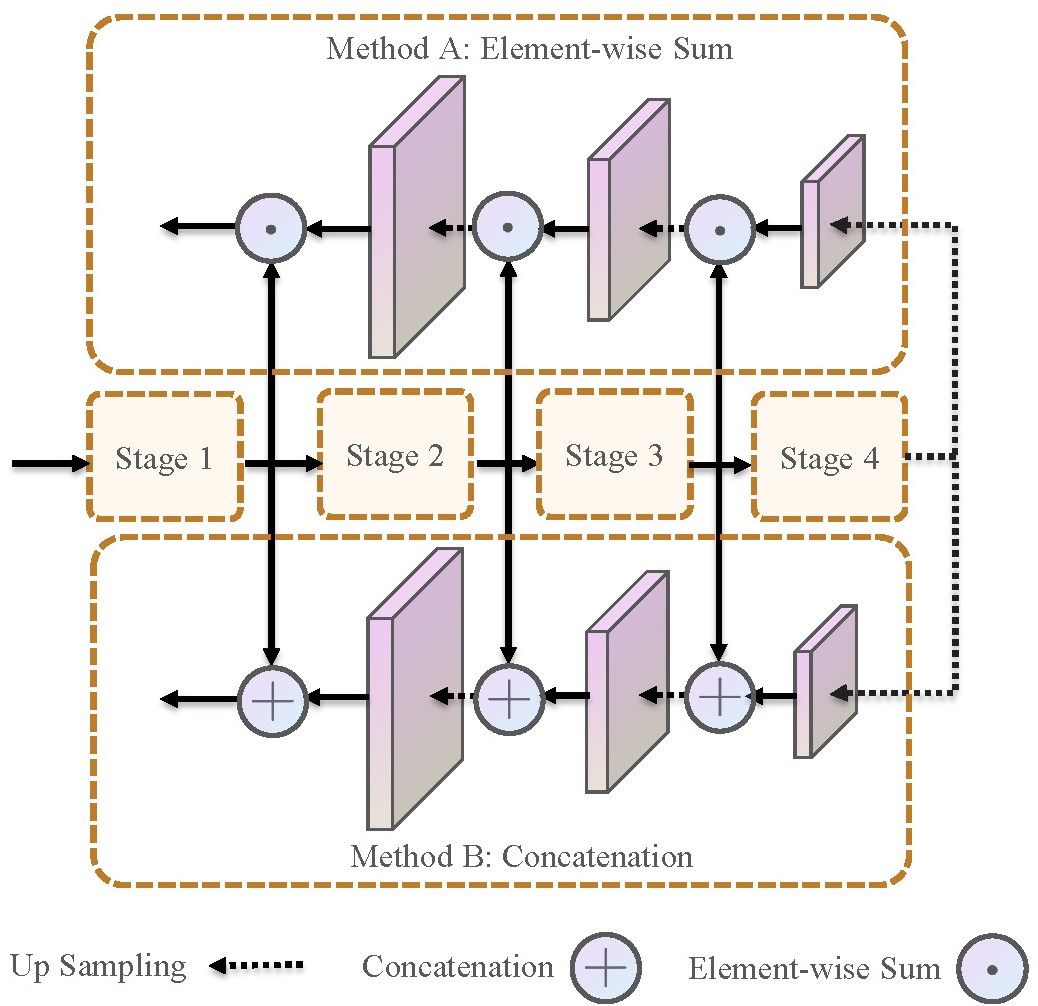}
    \caption{Feature Fusion Details.}
    \label{fig:concat and sum}
    \vspace{-0.2in}
\end{figure}

As shown in Fig. \ref{fig:concat and sum}, we adopt two different methods to examine the effects of multi-scale feature fusion. Method A is the element-wise sum. It generates the new features from the summation operation of the original features, but a little information of the original features will be lost during the process. Method B concatenates those feature maps by splicing each feature map directly along the direction of the channel, thus all information will be kept.

\begin{table*}[t!]
\centering
\scriptsize
\caption{Comparison on the COCO validation set.}
\label{tab:evaluation-result}
\resizebox{\linewidth}{!}{
\begin{tabular}{@{}l|l|c|c|c|c|cccccc@{}}
\toprule
Method & Backbone & Pretrain & Input size & \#Params & GFLOPs & $\mathrm{AP}$ & $\mathrm{AP}^{50}$ & $\mathrm{AP}^{75}$ & $\mathrm{AP}^{M}$ & $\mathrm{AP}^{L}$ & $\mathrm{AR}$ \\ \midrule
\begin{tabular}[c]{@{}l@{}}8-stage Hourglass\cite{newell2016stacked}  \\ CPN\cite{chen2018cascaded} \\ CPN + OHKM\cite{chen2018cascaded}  \\ SimpleBaseline\cite{xiao2018simple}  \\ SimpleBaseline\cite{xiao2018simple}  \\ SimpleBaseline\cite{xiao2018simple}\end{tabular} & \begin{tabular}[c]{@{}l@{}}8-stage Hourglass \\ ResNet-50 \\ ResNet-50 \\ ResNet-50 \\ ResNet-101 \\ ResNet-152\end{tabular} & \begin{tabular}[c]{@{}l@{}}N \\ Y \\ Y \\ Y \\ Y \\ Y\end{tabular} & \begin{tabular}[c]{@{}l@{}}256 × 192 \\ 256 × 192 \\ 256 × 192 \\ 256 × 192 \\ 256 × 192 \\ 256 × 192\end{tabular} & \begin{tabular}[c]{@{}l@{}}25.1M \\ 27.0M \\ 27.0M \\ 34.0M \\ 53.0M \\ 68.6M\end{tabular} & \begin{tabular}[c]{@{}l@{}}14.3 \\ 6.20 \\ 6.20 \\ 8.90 \\ 12.4 \\ 15.7\end{tabular} &  
\begin{tabular}[c]{@{}l@{}}66.9 \\ 68.6 \\ 69.4 \\ 70.4 \\ 71.4 \\ 72.0\end{tabular} &  
\begin{tabular}[c]{@{}c@{}}{--} \\ {--} \\ {--} \\ 88.6 \\ 89.3 \\ 89.3\end{tabular} &  
\begin{tabular}[c]{@{}c@{}}{--} \\ {--} \\ {--} \\ 78.3 \\ 79.3 \\ 79.8\end{tabular} &  
\begin{tabular}[c]{@{}c@{}}{--} \\ {--} \\ {--} \\ 67.1 \\ 68.1 \\ 68.7\end{tabular} &  
\begin{tabular}[c]{@{}c@{}}{--} \\ {--} \\ {--} \\ 77.2 \\ 78.1 \\ 78.9\end{tabular} &  
\begin{tabular}[c]{@{}c@{}}{--} \\ {--} \\ {--} \\ 76.3 \\ 77.1 \\ 77.8\end{tabular}   
\\
\midrule
\begin{tabular}[c]{@{}l@{}}HRNet-W32\cite{wang2020deep} \\ HRNet-W32\cite{wang2020deep} \\ HRNet-W48\cite{wang2020deep}\end{tabular} & \begin{tabular}[c]{@{}l@{}}HRNet-W32 \\ HRNet-W32 \\ HRNet-W48\end{tabular} & \begin{tabular}[c]{@{}l@{}}N \\ Y \\ Y\end{tabular} & \begin{tabular}[c]{@{}l@{}}256 × 192 \\ 256 × 192 \\ 256 × 192\end{tabular} & \begin{tabular}[c]{@{}l@{}}28.5M \\ 28.5M \\ 63.6M\end{tabular} & \begin{tabular}[c]{@{}l@{}}7.10 \\ 7.10 \\ 14.6\end{tabular} &  
\begin{tabular}[c]{@{}l@{}}73.4 \\ 74.4 \\ 75.1\end{tabular} &
\begin{tabular}[c]{@{}l@{}}89.5 \\ 90.5 \\ 90.6\end{tabular} &
\begin{tabular}[c]{@{}l@{}}80.7 \\ 81.9 \\ 82.2\end{tabular} &
\begin{tabular}[c]{@{}l@{}}70.2 \\ 70.8 \\ 71.5\end{tabular} &
\begin{tabular}[c]{@{}l@{}}80.1 \\ 81.0 \\ 81.8\end{tabular} &
\begin{tabular}[c]{@{}l@{}}78.9 \\ 79.8 \\ 80.4\end{tabular} 
\\ \midrule
\begin{tabular}[c]{@{}l@{}}SimpleBaseline\cite{xiao2018simple}  \\ HRNet-W32\cite{wang2020deep} \\ HRNet-W48\cite{wang2020deep}\end{tabular} & \begin{tabular}[c]{@{}l@{}}ResNet-152 \\ HRNet-W32 \\ HRNet-W48\end{tabular} & \begin{tabular}[c]{@{}l@{}}Y \\ Y \\ Y\end{tabular} & \begin{tabular}[c]{@{}l@{}}384 × 288 \\ 384 × 288 \\ 384 × 288\end{tabular} & \begin{tabular}[c]{@{}l@{}}68.6M \\ 28.5M \\ 63.6M\end{tabular} & \begin{tabular}[c]{@{}l@{}}35.6 \\ 16.0 \\ 32.9\end{tabular} &
\begin{tabular}[c]{@{}l@{}}74.3 \\ 75.8 \\ 76.3\end{tabular} &
\begin{tabular}[c]{@{}l@{}}89.6 \\ 90.6 \\ 90.8\end{tabular} &
\begin{tabular}[c]{@{}l@{}}81.1 \\ 82.7 \\ 82.9\end{tabular} &
\begin{tabular}[c]{@{}l@{}}70.5 \\ 71.9 \\ 72.3\end{tabular} &
\begin{tabular}[c]{@{}l@{}}79.7 \\ 82.8 \\ 83.4\end{tabular} &
\begin{tabular}[c]{@{}l@{}}79.7 \\ 81.0 \\ 81.2\end{tabular} \\ \midrule
\begin{tabular}[c]{@{}l@{}} \textbf{Ours} \end{tabular} & \begin{tabular}[c]{@{}l@{}} Swin-L with concatenation \end{tabular} & \begin{tabular}[c]{@{}l@{}} Y \end{tabular} & \begin{tabular}[c]{@{}l@{}} 384 × 384 \end{tabular} &  \begin{tabular}[c]{@{}l@{}} 197.9M\end{tabular} & \begin{tabular}[c]{@{}l@{}} 204.5\end{tabular} &
\begin{tabular}[c]{@{}l@{}} 76.2 \end{tabular} &
\begin{tabular}[c]{@{}l@{}} \textbf{93.4} \end{tabular} &
\begin{tabular}[c]{@{}l@{}} \textbf{83.3} \end{tabular} &
\begin{tabular}[c]{@{}l@{}} \textbf{72.6} \end{tabular} &
\begin{tabular}[c]{@{}l@{}} 81.5 \end{tabular} &
\begin{tabular}[c]{@{}l@{}} \textbf{84.7} \end{tabular} \\
\begin{tabular}[c]{@{}l@{}} \end{tabular} & \begin{tabular}[c]{@{}l@{}} Swin-L with element-wise sum \end{tabular} & \begin{tabular}[c]{@{}l@{}} Y \end{tabular} & \begin{tabular}[c]{@{}l@{}} 384 × 384 \end{tabular} &  \begin{tabular}[c]{@{}l@{}} 196.4M\end{tabular} & \begin{tabular}[c]{@{}l@{}} 202.6\end{tabular} &
\begin{tabular}[c]{@{}l@{}} \textbf{76.3} \end{tabular} &
\begin{tabular}[c]{@{}l@{}} \textbf{93.5} \end{tabular} &
\begin{tabular}[c]{@{}l@{}} \textbf{83.4} \end{tabular} &
\begin{tabular}[c]{@{}l@{}} \textbf{72.5} \end{tabular} &
\begin{tabular}[c]{@{}l@{}} 81.7 \end{tabular} &
\begin{tabular}[c]{@{}l@{}} \textbf{84.7} \end{tabular} \\
\bottomrule
\end{tabular}%
}
\end{table*}

\subsection{Analysis}
\vspace{-0.1in}
We think transformer structure is more suitable for human pose estimation tasks than convolutional neural networks, as CNNs use fixed-size filters to extract features from input images. Thus, they usually only have a fixed-sized reception field and lack the ability to gain relationships between far apart pixels. However, getting this long-range dependence information is essential to human pose estimation tasks. It can help locate the position of body joints that are not close to each other or even occluded. With the help of the attention mechanism, transformer calculates the similarity and relation between different pixels within the range and assign weights accordingly, thus gaining the ability to extract long-range dependence information.

But transformer models are generally larger than CNN models, and they require more training data to achieve a comparable result. Besides the model we trained with pre-trained weights, we also tried to train this transformer-based model from scratch to determine how many differences there would be. From the experiment results, the AP of the model trained from scratch can be as much as 20\% worse than that of the pretrained one and proves the importance of pretraining.

\section{Experiments}
\vspace{-0.1in}
\SubSection{Dataset}
\vspace{-0.1in}
The COCO person keypoints detection dataset\cite{lin2014microsoft} contains over 200,000 images and 250,000 person instances. In this dataset, the person instances are all labelled with 17 keypoints, consisting of 5 face landmarks and 12 body joints. All the keypoints are annotated with three values: the x coordinate, the y coordinate, and a flag indicating whether the keypoint is visible and labelled.  Most of the persons are at medium or large scale, and a large number of the persons are only partially visible, thus bringing challenges to our task. The training set train2017 contains more than 150,000 people, and the evaluation set val2017 contains 5,000 images. 

\SubSection{Evaluation Metric}
\vspace{-0.1in}
The main objective of human pose estimation is predicting the keypoints as close as possible to the ground truth. Thus a precise method to evaluate the prediction is needed. We adopt Object Keypoint Similarity (OKS)\cite{lin2014microsoft} to evaluate our model, which is inspired by object detection metrics, and commonly used to evaluate models in human pose estimation. It calculates the degree of match between predicted and ground truth, and normalized by the person scale. The result is within the range of 0 to 1, a higher value means the prediction is closer to the ground truth.



We report the average precision (AP) and average recall (AR) scores following the standard COCO format, where AP is calculated as the mean average precision at 10 positions when the value of OKS are 0.50, 0.55, 0.60, ..., 0.90, 0.95, and \(AP^{50}\), \(AP^{75}\), \(AP^{M}\),  \(AP^{L}\) are also recorded. 


\SubSection{Settings}
\vspace{-0.1in}
There are four Swin Transformers with different size, namely Swin Tiny (Swin-T), Swin Small (Swin-S), Swin Base (Swin-B), and Swin Large (Swin-L). In order to achieve the best result, we choose Swin-L here, with weights pretrained on ImageNet-22K dataset which contains 14.2 million images and 22K classes. The input image size is set to $384 \times 384$, the embedding dimension for the first stage is set to $C = 192$, the window size is set to 12, the number of blocks and number of heads for each stage are set to $\{2, 2, 18, 2\}$ and $\{6, 12, 24, 48\}$, respectively. The total model size and computation complexity is about $2\times$ the Swin-B model. We set the batch size to 10, which is limited by the memory, and set the number of epoches to 240. Adam optimizer is used, with initial learning rate experimentally set to 5e-5, and learning rate is reduced by a factor of 10 when the epoch reaches 60, 120 and 160.

\SubSection{Results}
\vspace{-0.1in}
We compare our results with previous state-of-the-art human pose estimation models, including SimpleBaseline, Cascaded Pyramid Network (CPN), Hourglass, and HRNet, and show the results in Table. \ref{tab:evaluation-result}. As it shows, our two models with concatenation and element-wise sum, respectively, both achieve a better result on $AP^{50}$, $AP^{75}$ and $AP^{M}$, and have 3.5\% improvement on $AR$. Besides that, Swin-L with element-wise sum achieves the same $AP$ as HRNet-W48.

\begin{table}[h]
\scriptsize
\centering
\caption{Results of different fusion methods, input resolution, and model size}
\label{tab:ablation-study}
\resizebox{\columnwidth}{!}{%
\begin{tabular}{@{}c|c|c|c|c|c|c|cc@{}}
\toprule
ID & Backbone & Pretrain\tnote{1} & Fusion Method\tnote{2} & Input size & \#Params & GFLOPs & \multicolumn{1}{l}{AP} & {AR} \\ \midrule
1 & Swin-S & 1K & Concat & 224 × 224 & 49.5M & 11.9 & 71.6 & 75.2\\ \midrule
2 & Swin-S & 1K & Sum & 224 × 224 & 49.1M & 11.7 & 71.7 & 75.2 \\ \midrule
3 & Swin-B & 22K & Concat & 224 × 224 & 88.0M & 15.9 & 72.6 & 75.9 \\ \midrule
4 & Swin-B & 22K & Sum & 224 × 224 & 87.3M & 15.7 & 72.2 & 75.7 \\ \midrule
5 & Swin-L & 22K & Concat & 224 × 224 & 197.9M & 24.2 & 72.8 & 76.1 \\ \midrule
6 & Swin-L & 22K & Concat & 384 × 384 & 197.9M & 204.5 & 76.2 & 84.7\\ \midrule
7 & Swin-L & 22K & Sum & 224 × 224 & 196.4M & 23.6 & 72.9 & 76.3 \\ \midrule
8 & Swin-L & 22K & Sum & 384 × 384 & 196.4M & 202.6 & 76.3 & 84.7\\ \bottomrule
\end{tabular}%
}
\begin{tablenotes}
\footnotesize
   \item[1] Pretrain: models pretrained on ImageNet-1K or ImageNet-22K. 
   \item[2] Fusion Method: Concat (Concatenation), Sum (Element-wise Sum)
   
\vspace{-0.2in}
\end{tablenotes}
\end{table}

\section{Ablation Study} 
\label{ablationstudy}
\vspace{-0.1in}
\SubSection{Input Resolution}
\vspace{-0.1in}
An ablation study is chosen to determine the factor of input resolution that affects the performance. The results of our model based on Swin-L are presented in Table. \ref{tab:ablation-study}, with two fusion methods, and input resolution set to $224 \times 224$ and $384 \times 384$, respectively. As shown in the table, comparing experiments 5, 6, 7 and 8, with the input resolution increased, the average precision has increased by 3.4 points for both fusion methods, with the price of a significant larger GFLOPs.  

\SubSection{Model Size}
\vspace{-0.1in}
The experiments were conducted over the different sizes of Swin-B and Swin-L. As shown in the Table. \ref{tab:ablation-study}, comparing the two results between experiments 3 and 5, it can be seen that the larger size of backbone only contributes limited improvements to its results. Similarly, no significant increase was found in experiment 4 compared with experiment 7. In addition, the computation cost and model size increase significantly, as the result of the growing FLOPs and number of parameters.
Overall, these results indicate the trade-offs between precision and computation cost. A smaller backbone should be selected if the computation cost is of higher priority.

\SubSection{Fusion Approach}
\vspace{-0.1in}
We study how two different fusion methods, namely element-wise sum and concatenation along channels, affect the performance of our models. It has been demonstrated that concatenation is able to retain as much information as possible during the fusion process. However, the number of parameters for concatenation will be greater than element-wise sum, which results in higher cost from the calculation. In particular, an independent ablation experiment was performed to determine whether there was a difference between the two fusion methods. 

In summary, it can be seen from the data in Table. \ref{tab:ablation-study}, these results show that the computation cost of concatenation is slightly higher than the element-wise sum.

\section{Conclusion}
\vspace{-0.1in}
In this paper, we have explored the application of transformer to human pose estimation and presented two slightly different versions of models. 
The hierarchical structure of the Swin Transformer inspires us to adopt feature pyramid, fuse the feature maps output by different stages together, and build a more substantial model. The two proposed models achieve competitive results compared to the famous HRNet-W48 and other CNN based models.

Although the results demonstrate that transformer models can be successfully implemented to human pose estimation tasks with competitive performance, the main challenge is that transformer models are huge with almost two times more parameters and six times more FLOPs. Further studies need to be carried out in order to reduce the computation cost and obtain an outstanding performance compared with CNN models.

\nocite{toshev2014deeppose,chu2016structured,li2020human,wang2021lower,wei2016convolutional,vaswani2017attention,dosovitskiy2020image,liu2021swin,lin2014microsoft,felzenszwalb2009object,deng2020feature,kong2016hypernet,he2015spatial,krizhevsky2012imagenet,xiao2018simple,wang2020deep,he2016deep,yang2021transpose}
\bibliographystyle{latex8}
\bibliography{latex8}

\end{document}